%% file: example_paper.tex
\documentclass{article}

\usepackage{microtype}
\usepackage{graphicx}
\usepackage{subfigure}
\usepackage{booktabs} 

\usepackage{hyperref}
\usepackage{amsmath}

\DeclareMathOperator*{\argmin}{arg\,min}

\usepackage[accepted]{icml2024}

\usepackage{amsmath}
\usepackage{amssymb}
\usepackage{mathtools}
\usepackage{amsthm}
\usepackage{enumitem}
\usepackage{multirow}

\usepackage[capitalize,noabbrev]{cleveref}

\theoremstyle{plain}
\newtheorem{theorem}{Theorem}[section]
\newtheorem{hypothesis}[theorem]{Hypothesis}
\newtheorem{proposition}[theorem]{Proposition}

\theoremstyle{definition}

\theoremstyle{remark}

\usepackage[textsize=tiny]{todonotes}

\icmltitlerunning{Peeking Behind the Curtains of Residual Learning}

\begin{document}

\DeclarePairedDelimiter\ceil{\lceil}{\rceil}
\DeclarePairedDelimiter\floor{\lfloor}{\rfloor}

\twocolumn[
\icmltitle{Peeking Behind the Curtains of Residual Learning}




\begin{icmlauthorlist}
\icmlauthor{Tunhou Zhang}{yyy}
\icmlauthor{Feng Yan}{zzz}
\icmlauthor{Hai Li}{yyy}
\icmlauthor{Yiran Chen}{yyy}
\end{icmlauthorlist}

\icmlaffiliation{yyy}{Department of Electrical and Computer Engineering, Duke University, Durham, NC}
\icmlaffiliation{zzz}{Department of Electrical and Computer Engineering, University of Houston, Houston, TX}

\icmlcorrespondingauthor{Yiran Chen}{yiran.chen@duke.edu}

\icmlkeywords{Machine Learning, ICML}

\vskip 0.3in
]



\printAffiliationsAndNotice{}  

\begin{abstract}
The utilization of residual learning has become widespread in deep and scalable neural nets.
However, the fundamental principles that contribute to the success of residual learning remain elusive, thus hindering effective training of plain nets with depth scalability.
In this paper, we peek behind the curtains of residual learning by uncovering the ``dissipating inputs'' phenomenon that leads to convergence failure in plain neural nets: the input is gradually compromised through plain layers due to non-linearities, resulting in challenges of learning feature representations.
We theoretically demonstrate how plain neural nets degenerate the input to random noise and emphasize the significance of a residual connection that maintains a better lower bound of surviving neurons as a solution.
With our theoretical discoveries, we propose ``The Plain Neural Net Hypothesis'' (PNNH) that identifies the internal path across non-linear layers as the most critical part in residual learning, and establishes a paradigm to support the training of deep plain neural nets devoid of residual connections. 
We thoroughly evaluate PNNH-enabled CNN architectures and Transformers on popular vision benchmarks, showing on-par accuracy, up to $0.3\times$ higher training throughput, and 2$\times$ better parameter efficiency compared to ResNets and vision Transformers.
\end{abstract}

\input{_txt/1_Introduction}
\input{_txt/2_Related_Work}
\input{_txt/3_Residual_Learning}

\input{_txt/4_ConvCoder}
\input{_txt/5_Experiments}

\input{_txt/7_Conclusion}

\bibliography{references}
\bibliographystyle{icml2024}

\newpage
\appendix
\input{_txt/8_Appendix}
\onecolumn


\end{document}

%% file: _txt/1_Introduction.tex
\section{Introduction}
Deep neural networks (DNNs) have been instrumental in the significant advances witnessed in vision recognition applications, including classification~\cite{deng2009imagenet}, detection~\cite{lin2014microsoft}, and segmentation~\cite{Cordts2016Cityscapes}. 
As a research foundation, the VGG model~\cite{simonyan2014very} has exemplified a harmonious balance between performance and hardware efficiency by constructing a moderate depth-principled hierarchical architecture.
The emergence of residual learning in ResNet later facilitates the development of a considerably deeper Convolutional Neural Net (CNN), thus stimulating the trend of ``deep learning'' and establishing ``depth'' as a defining characteristic of ConvNets. Subsequently, this deep design language has inspired a proliferation of models, incorporating various manual design philosophies~\cite{xie2017aggregated,huang2017densely,szegedy2017inception}, as well as automatic patterns~\cite{zoph2018learning,xie2019exploring}. These diverse models have further refined neural nets, especially CNNs.
Each of these existing works demonstrates a unique emphasis on aspects such as performance, efficiency, and scalability.

To elucidate the theoretical underpinnings of residual learning, a suite of theories and insights, encompassing both data-dependent approaches~\cite{balduzzi2017shattered,xiao2018dynamical} and data-agnostic methodologies~\cite{zagoruyko2017diracnets,ding2021repvgg,meng2021rmnet}, have been proposed.
Existing data-dependent methods~\cite{li2020residual} focus primarily on the transfer of knowledge from a pre-trained ResNet towards a plain neural net. 
However, this strategy culminates in sub-par depth scalability, as the intrinsic challenges associated with learning deep plain representations persist unmitigated.
Existing data-agnostic approaches focus on analyzing the forward and/or backward propagation process during ResNet training. These works attempt to simulate such behavior in plain CNNs through sophisticated engineering efforts, such as re-parameterization~\cite{ding2021repvgg}, over-parameterization~\cite{meng2021rmnet}, and local training~\cite{wang2021revisiting}. However, these approaches may lead to poor training efficiency. For example, RepVGG~\cite{ding2021repvgg} involves 2$\times$ training overheads with at least 1.4$\times$ larger network size.

\begin{figure*}[t]
\begin{center}
    \includegraphics[width=0.8\linewidth]{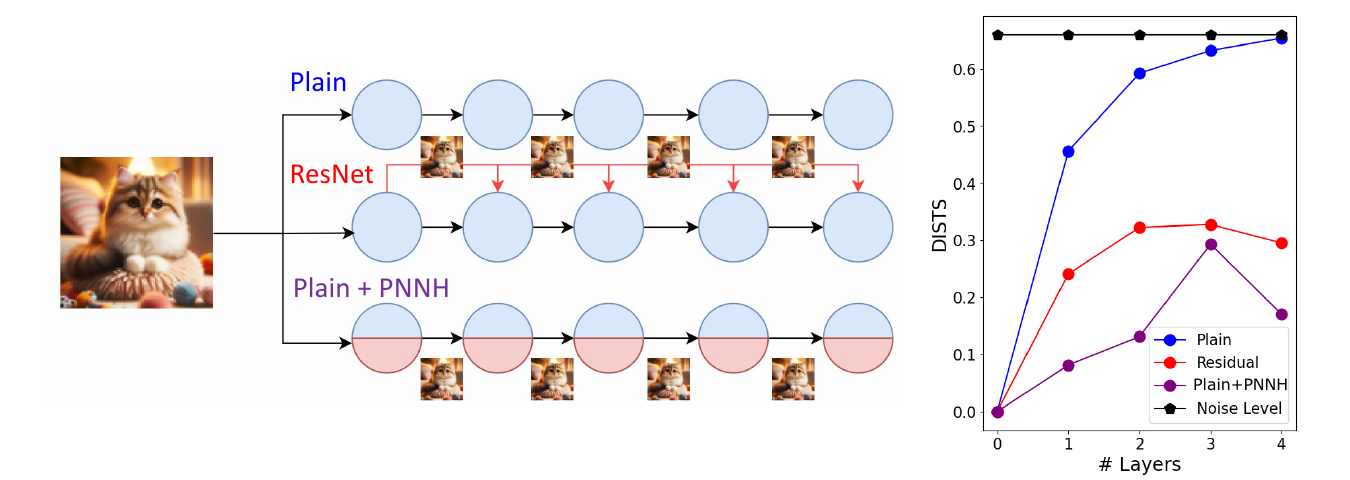}
    \vspace{-1em}
     \caption{While plain CNNs does not connect features from source to deeper layers, the input information is gradually lost as is reflected from dissimilarity score (DISTS), causing ``dissipating input''. ResNet addresses ``dissipating input'' by establishing an explicit residual connection, and our proposed PNNH paradigm derives theoretical foundations to incorporate residuals into plain architecture.}
    \label{fig:intro_fig}
    \vspace{-2em}
\end{center}
\end{figure*}
In this work, we make an initial effort to provide a formal explanation of residual learning. With weight parameters at He-normal initialization~\cite{he2016deep}, we uncover the ``dissipating input'' phenomenon in deep plain CNNs, see Figure \ref{fig:intro_fig}. 
Specifically, we argue that plain CNNs cannot preserve critical input information and quickly degenerate the input information to random noise: the DISTS dissimilarity~\cite{ding2020image} between VGG layer-wise outputs and original inputs quickly approaches the noise level after only five plain layers.
As a result, the layer output loses its relationship with the original input and thus fails to converge to correct representations in plain networks.

We formulate theoretical foundations to verify our discovery and further articulate ``the Plain Neural Net Hypothesis''(PNNH) to pioneer the possibility of training plain CNNs from scratch.
PNNH states the importance of an internal path within neural nets to preserve cross-layer input information.
We establish the "PNNH paradigm" to practice PNNH on both CNN architectures and Transformer models for vision problems.
Within the PNNH paradigm, we split a conventional neural net layer (e.g., a CNN layer) into two parts: a \textbf{learner} that learns the representation of data-dependent features and a \textbf{ coder} that transforms the preceding input features into efficient coding.
While the representer learns knowledge from input data via SGD, the coder preserves input information for better generalization, with weights shared across different neural net layers to provide better parameter efficiency for CNN and Transformer architectures.
The combination of a learner and a coder provides a simplified training procedure, leading to improved training efficiency without residuals.

We apply PNNH on both CNN and Transformer architectures and evaluate the performance on vision benchmarks. 
On CNN architectures, we prototype PNNH on three popular CNN blocks: vanilla residual block~\cite{he2016deep}, residual bottleneck block~\cite{he2016deep}, inverted residual bottleneck block~\cite{sandler2018mobilenetv2}.
In Transformer, we prototype PNNH on the residual MLP block of the Swin Transformer~\cite{liu2021swin}.
We conduct extensive experiments on three vision classification benchmarks: CIFAR-10, CIFAR-100, and ImageNet-1K.
With light tuning on hyperparameters, PNNH allows plain CNN/Transformer architectures to achieve on-par results with the ResNet/Transformer counterparts, achieving up to $0.3\times$ higher training throughput and 2$\times$ parameter efficiency.
We summarize our contributions as follows.
\begin{itemize}[noitemsep,leftmargin=*]
    \item Our work peeks behind the curtains of residual learning by explaining the ``dissipating inputs'' behaviors of plain CNN learning. We theoretically verify our discovery by deriving $\delta$-$\epsilon$ bound on the minimum number of layers that collapse a plain/residual CNN and propose ``the Plain Neural Net Hypothesis''(PNNH) to mitigate ``dissipating inputs''.
    \item We introduce a novel paradigm to practice PNNH on plain neural nets. 
    Within the paradigm, PNNH adopts a combination of the learner and the coder to establish an internal path within plain neural nets. PNNH enables a plain network to be trained from scratch in the same fashion as residual neural nets, without additional engineering efforts such as tuning, distillation, and reparameterization. 
    \item We thoroughly evaluate PNNH on CNN/Transformer architectures on three vision benchmarks and demonstrate promising signals in terms of performance, efficiency, and the underlying theoretical foundations.
\end{itemize}

%% file: _txt/2_Related_Work.tex
\section{Related Work}
\noindent \textbf{Residual Learning.} ResNet achieves great success in deep neural nets with a simple and elegant residual connection. However, the underlying reasons behind the success of residual learning have not yet been revealed. 
Previous work attributes the success of ResNet to easier optimization~\cite{he2016deep}, a group of shorter effective networks~\cite{veit2016residual}, increased model cardinality~\cite{xie2017aggregated}, and reduced shattered gradient phenomenon~\cite{balduzzi2017shattered}.
Despite the good insights provided by these works from theoretical analysis and empirical evaluations, these interpretations may not be applied to improve the generalization of plain CNNs from scratch.
Our work demonstrates that residual learning can maintain input information that otherwise would be gradually lost when fed into the consequent nonlinear neural net layers.
Based on this discovery, plain CNNs can achieve similar results as residual learning, since the input information can be preserved.

\noindent \textbf{Plain Deep Neural Nets.}
The rise of plain neural nets begins with the prevalence of heavily optimized VGG-style kernels~\cite{lavin2016fast}.
Existing work uses weight reparameterization~\cite{zagoruyko2017diracnets}, weight initialization~\cite{xiao2018dynamical}, knowledge distillation~\cite{li2020residual} to create convolutional neural nets without residual connections. Recent works~\cite{ding2021repvgg} also combine reparameterization and overparameterization techniques to enhance the representation of plain CNNs. Yet, these approaches may not provide direct solutions for plain CNNs to achieve similar performance benefits of residual learning via good depth scalability.
More recent work~\cite{meng2021rmnet} has produced an equivalent form of ResNet in a plain convolutional neural net with additional parameters and/or suboptimal performance.
In comparison, we exercise our theoretical understanding of residual learning (PNNH) and craft a deep plain neural net from scratch without any explicit form of residual connections.
As a result, plain CNN/transformer architectures with PNNH enabled better efficiency without comprising performance compared to the residual learning counterparts.

%% file: _txt/3_Residual_Learning.tex
\section{Why deep plain neural nets is challenging?}
The simple residual connection in ResNet pushes the success of deep residual neural networks.
A set of theoretical and empirical hypotheses attempts to reveal the secrets of residual learning in ResNets.
Yet, none of these studies reveal the explicit motifs of residual connections that lead to its success.
In this section, we start by introducing the ``dissipating inputs'' phenomenon we discovered in deep plain neural nets, and then proceed with formulating an exact hypothesis
``The Plain Neural Net Hypothesis'' (PNNH) to unravel the secrets in building deep plain neural nets. 
We summarize all useful annotations in Appendix.

\textit{Our key insight is that ReLU non-linearity drops negative neuron responses that contain the original information of the source input and establishes a lower bound of positive neurons survived from ReLU non-linearity in both plain VGG-style layers and residual ResNet-style layers.
The lower bounds demonstrate that a residual connection pushes up the mean of neuron responses and yields a higher portion of non-zero activations surviving the ReLU non-linearity.}
Next, we explain the scope and introduce the assumptions in our analysis.

\noindent \textbf{\textbf{Scope.}}
In this paper, we discuss convolutional neural nets and MLP neural nets within Transformer architectures. Without loss of generality, \textit{We limit the discussion to vanilla residual blocks with 2 3$\times$3 convolution layers to conduct the analysis and discuss the generality of broader choices of CNN/MLP block in the end.}
We focus on the weight initialization phase of neural nets, as it is critical to starting representation learning.
In this case, the network input $X$ is independent of the randomly initialized weights $W_{conv1}$ and $W_{conv2}$.
To simplify the analysis, we drop Batch Normalization (BN) because they are data-dependent. The behavior of BN can be fused into weights equivalently.
Similarly, we exclude bias coefficients in each layer.

\noindent \textbf{Assumptions.}
We assume that the input is non-negative activations with mean $\mu_{x}$ and variance $\sigma^2_{x}$.
We also assume $\sigma_{x}=o(\mu_{x}) > 0$. Without losing generality, we can always scale a non-negative input to the above scenario by linear operators.
Note that all our analyses are based on the He-normal initialization~\cite{he2016deep} of weight parameters, which is commonly employed in ResNets. Specifically, the weights in each layer $W \in \mathrm{R}^{C_{out} \times C_{in} \times K \times K}$ are initialized from a normal distribution as follows: 
\begin{equation}
{
    W \sim \mathcal{N}(0, \sqrt{2 / C_{in} \times K \times K}) ,}
    \label{eq:weight_init}
\end{equation}
where $C_{in}$ is the input channel number, $C_{out}$ is the output channel number, and $K$ is the kernel size.

\textbf{Remarks.} The ``$fan\_in$'' mode in He normal~\cite{he2016deep} is the most straightforward way to preserve the variance of the inputs $X$ when passing through convolutional doublets. 
Specifically, it doubles the variance of positive neurons prior to ReLU, and restores the variance following the ReLU non-linearity.
Our analysis also fits the scenario under other weight initializations, such as Glorot normal~\cite{glorot2010understanding}. This is because modern ConvNets have adopted batch normalization layers, and it can achieve the transition of scale towards other weight initialization towards He-normal via a learned affine layer.

\subsection{Dissipating Inputs Are Lost Activations}
We first look at the transformation of input structures via a simple Convolution-ReLU doublet. We demonstrate that ReLU non-linearity drops information in neural nets.

\begin{proposition}
(\textbf{Curse of Non-linearity).}
ReLU non-linearity: $ReLU(x)=\max(0, x)$ is a low-rank operator that induces information loss in neuron responses.
We define \textbf{``dissipating inputs''} as losing 1-$\epsilon$ of the original information from negative neurons that are zeroed by ReLU non-linearity.
\label{proposition:curse_non_linear}
\end{proposition}

This is because a linear finite-width neural network layer cannot learn an unbiased estimator of the ReLU non-linearity. 

\textbf{Remarks.} 
Information loss induced by negative activations in ``dissipating inputs'' cannot be restored by the following layers initialized independent of data. In this aspect, leaky ReLU preserves more information than ReLU and thus encourages convergence in very deep plain networks. We empirically validate this proposition with experiments and detailed discussion in the Appendix.

Unlike the existing analysis of vanishing gradients~\cite{li2020residual} that focuses on the back propagation of convolutional neural nets, ``dissipating inputs'' focuses on forward propagation that emphasizes the initial stage of representation learning.
In this regard, it is important to study the probability of maintaining positive activations from an arbitrary input. With He-normal initialization which keeps the variance scaled, we first study the distribution of output neuron responses before the last ReLU non-linearity, as follows.
\vspace{-0.5em}
\begin{theorem}
\textbf{(Statistics of Neuron Responses).}
Before the last ReLU non-linearity, the output neuron responses in plain layers has zero-mean and $2\sigma_{x}^2$ variance. The use of a residual connection makes the output neuron responses have $\mu_{x}$ mean and $3\sigma_{x}^2$ variance.
\end{theorem}
\vspace{-0.5em}
\noindent \textbf{Remarks.} We utilize the variance-preserving mechanism in He-normal initialization in derivation. 

With the mean and variance of neurons calculated, We derive the lower bounds for maintaining positive activations in a plain layer and a residual layer as follows.

\begin{theorem}
\textbf{(Lower-bound on Surviving Neurons).} Given an arbitrary input with mean $\mu_{x}$ and variance $\sigma^2_{x}$, the probability that a neuron survives ReLU activations is lower bounded by $\frac{\delta^2}{4\sigma_{x}^2+{\delta}^2}$ in plain layers. For residual layers, the probability of maintaining positive activations is lower bounded by $\frac{(\mu_{x}+\delta)^2}{9\sigma_{x}^2 + {(\mu_{x}+\delta)}^2}$. Here, $\delta=o(\sigma_{x})$  determines the precision of the bound.
\label{th:surviving_neurons}
\end{theorem}

\vspace{-0.5em}

\noindent \textbf{Remarks.} We derive this bound by applying the Chebyshev-Cantelli inequality to the upper bound of the negative-side-tail probability distribution centered with 0 for plain layers and $\mu_{x}$ for residual layers.

With a total of $N$ plain/residual networks and $\epsilon$ of minimum surviving neurons to retrieve input, the minimum layers that collapse a plain/residual neural net design is given by the following $\delta-\epsilon$ bound:
\vspace{-0.5em}
\begin{equation}
{
    \inf[N_{plain}(\delta, \epsilon)] =\ceil{ \frac{\ln{(1/\epsilon)}}{\ln{[1 + (\frac{2\sigma_{x}}{\delta})^2]}}},
}
\label{eq:plain_layer_bound}
\end{equation}

\vspace{-0.5em}
\begin{equation}
    {
    \inf[N_{residual}(\delta, \epsilon)] = \ceil{ \frac{\ln{(1/\epsilon)}} {\ln{[1 + (\frac{3\sigma_{x}}{ \mu_{x}+\delta})^2]}}}.
    }
    \label{eq:residual_layer_bound}
\end{equation}
\vspace{-0.5em}

The lower bounds for plain layers and residual layers provide an important insight for the ResNet designs: pushing the mean of neuron distributions far from zero mean such that negative responses occur less often prior to ReLU activation. 
With $\sigma_{x}=o(\mu_{x})$, residual learning always achieves a better lower bound than plain learning shown in Theorem \ref{th:surviving_neurons}, thus allowing very deep architectures without losing significant amounts of information.
Furthermore, the bound in Theorem \ref{th:surviving_neurons} matches the observations in FixUp initialization~\cite{zhang2019fixup}, where a bias can serve an equivalent purpose as a residual connection to restore the representation power after zero-mean normalization layers.
As a result, the deep architecture can extract information from inputs, thanks to the preserved input information induced by the residual connections. Nevertheless, Eq. \ref{eq:residual_layer_bound} shows that very deep residual networks may still lose all input information and suffer worse generalization than their shallower counterparts.

For deeper convolutional neural net blocks, such as residual bottleneck block, the minimum layers that collapse a plain network should not be lower than $\inf[N_{plain}(\delta, \epsilon)]$ in Eq. \ref{eq:plain_layer_bound}. This is because an additional convolution layer with/without ReLU activation retrieves no input information, thus cannot increase the lower bound on surviving neurons. Thus, Eq. \ref{eq:plain_layer_bound} and Eq. \ref{eq:residual_layer_bound} serves as a lower bound for deeper convolutional neural net blocks, such as residual/inverted residual bottleneck blocks.

\subsection{The Plain Neural Net Hypothesis}

From a theoretical perspective, the key to residual learning success lies in establishing an internal path that maintains the input structures.
Theoretically, the residual connection uses positive-centered activations to push the zero-centered neuron responses further toward positive values, and thus improves the bounds in Theorem \ref{th:surviving_neurons} compared to their plain counterparts. 
Based on these studies, we propose below ``The Plain Neural Net Hypothesis'' that enables training of plain neural nets from scratch without extra engineering efforts, such as tuning, reparameterization, and distillation.
\begin{hypothesis}
\textbf{(The Plain Neural Net Hypothesis)}. A Deep plain CNN is trainable from scratch, as long as there exists an internal path that communicates input information towards output layer responses prior to non-linearity in every plain neural net layer.
\label{hypothesis:plain}
\end{hypothesis}

In particular, a residual connection is a special type of internal path represented by a full-rank identity mapping. However, a residual connection consumes more memory than the plain counterparts.
In the following section, we incorporate PNNH into plain neural nets and establish a simple yet novel paradigm to enable high-quality and convenient training of plain neural nets from scratch.

%% file: _txt/4_ConvCoder.tex
\section{Practicing PNNH on Plain Architectures}
\begin{figure*}[t]
\begin{center}
    \includegraphics[width=0.8\linewidth]{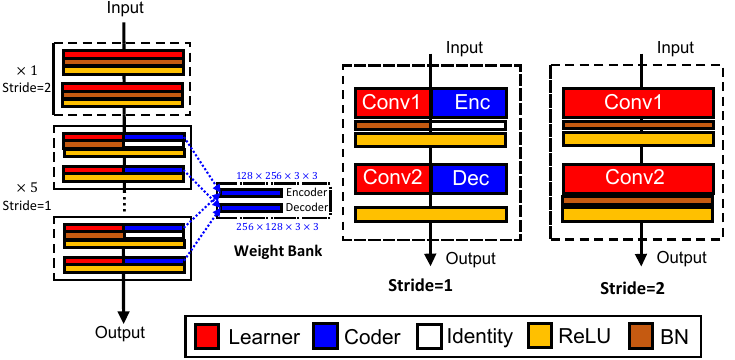}
    \vspace{-1em} 
    \caption{An example of applying PNNH on the 3rd stage of ResNet-34. We maintain the original block in the ``stride=2'' block, and incorporate the PNNH mechanism to ``stride=1'' blocks. }
    \label{fig:PNNH}
    \vspace{-2em}
\end{center}
\end{figure*}

We apply the aforementioned discoveries in PNNH to plain CNN / transformer architectures to establish a new paradigm, dubbed "PNNH paradigm", to mitigate the phenomenon of ``dissipating inputs'', enabling the training of very deep plain neural nets from scratch.
Figure \ref{fig:PNNH} illustrates an example of PNNH paradigm on a 34-layer plain neural net, following the same layer organization as ResNet-34. 
Within each layer of the plain neural net, PNNH establishes a dual-path structure within a plain neural net containing a \textbf{learner} and \textbf{coder}.
The learner enables supervised learning from data, and the coder employs an unsupervised auto-encoder to maintain input via an established internal path. 
The weights of the coder within each stage are shared to improve the parameter efficiency and training efficiency of PNNH.
Next, we elaborate on the coder design and the overall training procedure of the PNNH paradigm.


\subsection{Weight-Sharing Coder as Internal Path}
Within the coder design, we establish an internal path using an autoencoder~\cite{vincent2010stacked} as the coder model to maintain efficient coding of input structures.
Given $C_{in}$ input channels, the coder: $f_{coder}: X\in \mathrm{R}^{C_{in} \times K\times K} \to \hat{X}\in \mathrm{R}^{C_{in} \times K\times K}$ only maintains $C_{B}$ channels of efficient coding.
The goal of the coder is to reconstruct the input to minimize coder loss $l_{coder}$ as follows:
\begin{equation}
{
    L_{coder}(x|f_{coder}) = ||x-f_{coder}(x))||_2^{2} .
    \label{eq:ae_loss}
}
\end{equation}
\vspace{-1em}

We utilize weight-sharing that ties the encoder/decoder weights of each coder structure to the same weight bank. This is because the coder serves the same functionality (i.e., preserve input information) in each CNN block. Once the coder is trained, it is not updated with incoming input data. Thus, it is not necessary to train a coder uniquely for each CNN block.

Next, we discuss the detailed coder configurations that adapt towards different CNN blocks, including vanilla ResNet block, ResNet bottleneck block, and Inverted Residual Bottleneck block. This showcases the broad use of PNNH on various CNN architectures, such as vanilla ResNet (e.g., ResNet-18/34), bottleneck ResNet (e.g., ResNet-50/101), and inverted residuals (e.g., MobileNet-V2). We first illustrate the representation of a residual connection in a coder.


\noindent \textbf{A Residual Connection is a Special Coder.}
Following Eq. \ref{eq:ae_loss}, the coder contains the solution space for spatial and local feature learning.
More importantly, (partially) residual connections proposed in RM operation~\cite{meng2021rmnet} is a special case in the solution space of coder -- the encoder weights $W_{enc}\in \mathbb{R}^{ C_{B} \times C_{out} \times K \times K}$ and decoder weights $W_{dec}\in \mathbb{R}^{C_{out} \times C_{B}  \times K \times K}$ are constructed as follows:
\vspace{-1em}
\begin{equation}
    W_{enc}^{(i, j, x, y)}, W_{dec}^{(i, j, x, y)} = \begin{cases}
    1, \text{if } i=j, x=y=1 \\
    0. \text{Otherwise}
    \end{cases},
    \label{eq:aeinit}
\end{equation}
where $i, j$ denotes the channel index, and $x, y$ denotes the filter index. Although a residual connection perfectly preserves the spatial input features, it randomly drops channel information to compromise the input structure. Furthermore, although residual connection (or RM operation) is an optimal solution in the theories of linear auto-encoder, that is, $W_{enc}^{T}W_{dec}=I$, the nonlinearity activation ReLU within the auto-encoder structure no longer guarantees the optimality of the above solution. Therefore, the optimization in the coder is necessary to better address the phenomenon of ``dissipating input''.

Next, we illustrate the coder function in vanilla residual blocks, residual bottleneck blocks, and inverted residual bottleneck blocks.
\vspace{-0.5em}
\begin{itemize}[noitemsep,leftmargin=*]
    \item \textbf{Vanilla Residual Block.} A vanilla residual block contains 2 3$\times$3 convolution. Hence, the coder function $f_{coder}$ is defined as:
    \vspace{-1em}
    \begin{equation}
        f_{coder} = ReLU[(x \circledast W_{enc})] \circledast W_{dec}.
    \label{eq:vanilla_coder_func}
    \end{equation}
    Given $C$ input channels, the coder function produces efficient coding with $C_{B}$ channels, and expands it back to the original dimension. In this work, we choose $C_B=C/2$ for simplicity.
    \item \textbf{Residual Bottleneck Block.} A residual bottleneck block first squeezes the input feature to a lower dimension $C_{low}$, extracts the features from the squeezed feature and expands the squeezed feature back to the original dimension. Therefore, the coder function $f_{coder}$ is defined as:
    \begin{equation}
        f_{coder} = ReLU[ReLU[(x \circledast W_{enc})] \circledast I_{3\times 3}]\circledast W_{dec}.         
    \label{eq:rb_coder_func}
    \end{equation}
    Where $I_{3\times3} \in \mathbb{R}^{C_{B} \times C_{B} \times K \times K}$ can be initialized in the same way as in Eq. \ref{eq:aeinit}. Similarly to the previous case, we choose $C_{B}=C_{low}/2$ to preserve half of the information in compressed bottlenecks for simplicity.
    \item \textbf{Inverted Residual Bottleneck Block.} An inverted bottleneck block first expands the input feature to a higher dimension $C_{high}$, then carries a depth-wise convolution to extract features and finally projects the output back to the original dimension. In this work, we choose $C_{B}=C/2$ for simplicity. Hence, the coder function $f_{coder}$ is defined as:
    \begin{equation}
        f_{coder} = ReLU[ReLU[(x \circledast W_{enc})] \hat{\circledast} I^{dw}_{3\times 3}]\circledast W_{dec}.         
    \label{eq:irb_coder_func}
    \end{equation}
    Here, $\hat{\circledast}$ denotes a depthwise convolution, We initialize depthwise kernel $I^{dw}_{3\times 3} \in \mathbb{R}^{C_{B} \times 1 \times K \times K}$ as:
    \begin{equation}
    I^{dw (., 1, x, y)}_{3\times 3} = \begin{cases}
    1, \text{if } x=y=1 \\
    0. \text{Otherwise}
    \end{cases},
    \label{eq:irb_coder_init}
\end{equation}
where $i, j$ denotes the channel index, and $x, y$ denotes the filter index.
    \item \textbf{Transformer MLP.} Transfomers such as ViT~\cite{dosovitskiy2020image}, Swin Transformer~\cite{liu2021swin} encode residual layers within MLP to enable deep architectures. Such setting can be viewed as a special case of the vanilla residual block with kernel size fixed to 1.
\end{itemize}

We establish an unsupervised task for the coder to learn an internal path that better preserves input.
For better generalization, we use a rectified normal distribution instead: $X \sim max(0, N(0, 1))$. 
This is because the distribution of CNN responses is mostly Gaussian, and ReLU non-linearity forces the input to be nonnegative.
We train the coder in the aforementioned task to minimize lost information.
The training objective can be achieved by optimizing coder weights $(W_{enc}, W_{dec})$ and obtain the optimal coder $f^{*}_{coder}$ as follows:
\begin{equation}
\small{}{
    f_{coder}^{*} = \argmin_{f_{coder}} 
    \mathop{\mathbb{E}}_{x \sim \max(0, \mathrm{N}(0, 1))} L_{coder}(x|f_{coder}).
}
\label{eq:ae_objective}
\end{equation}

The objective in Eq. \ref{eq:ae_objective} can be easily optimized via SGD~\cite{ruder2016overview}.

\subsection{Training a Plain CNN with PNNH}
\label{sec:train_convcoder}
We present the algorithm for training a plain neural net enabled by PNNH in Algorithm~\ref{alg:train_convcoder}. Here, we use the vanilla residual block (e.g., ResNet-34) as an example when practicing PNNH on CNN architectures.
Algorithm~\ref{alg:train_convcoder} simply constructs an internal path (i.e., $W_{coder}$) within each layer of the plain CNN architecture, and then freezes the internal path to learn useful representations via updating $W_{learner}$.
\newcommand{\pluseq}{\mathrel{{+}{=}}}
\newcommand{\minuseq}{\mathrel{{-}{=}}}

\renewcommand{\algorithmicrequire}{ \textbf{Input:}}
\renewcommand{\algorithmicensure}{ \textbf{Output:}}
\begin{algorithm}[h]
\small{}{
$\left. \textbf{Input:} \right.$ \newline
Set of CNN block channel sizes $\mathbf{C}$, Coder Objective $L_{coder}$, Coder function $f_{coder}$, learner objective $L_{learner}$, learner function $f_{learner}$, learner target dataset $\mathrm{D}$ \newline
$\left. \textbf{Hyperparameter:} \right.$ \newline
Coder learning rate $\alpha_{coder}$, Coder training steps $T_{coder}$, learner learning rate $\alpha_{learner}$, learner training steps $T_{learner}$. \newline
\textbf{begin}
\begin{algorithmic}
\FOR {$C_{in} \in \mathbf{C}$} 
    \STATE Initialize $W_{coder}=(W_{enc}, W_{dec}) \sim \mathbb{N}(0, 0.01)$
    \FOR {$t \gets 1$ to $T_{coder}$}
        \STATE $W_{coder} \minuseq \alpha_{coder}\nabla \mathop{\mathbb{E}}_{x \sim max(0, \mathbb{N}(0, 1))}L_{coder}(x, f_{coder})$
    \ENDFOR
    \STATE Copy $W_{coder}$ to all blocks with CNN containing $C_{in}$ channels.
\ENDFOR 
\STATE Initialize all learner: $W_{learner}=(W_{conv1}, W_{conv2})$ following He-normal initialization.

\FOR {$t \gets 1$ to $T_{learner}$}
    \STATE $W_{learner} \minuseq \alpha_{learner}\nabla \mathop{\mathbb{E}}_{x \in D} L_{learner}(x|f_{learner})$
\ENDFOR
\end{algorithmic}
\textbf{end}
}
\caption{Training a Plain CNN with PNNH.}
\label{alg:train_convcoder}
\end{algorithm}

%% file: _txt/5_Experiments.tex
\section{Experiments}
In this section, we practice ``The Plain Neural Net Hypothesis'' (PNNH) on plain architectures over three vision benchmarks: CIFAR-10~\cite{krizhevsky2009learning}, CIFAR-100~\cite{krizhevsky2009learning}, and ImageNet~\cite{deng2009imagenet}. We support PNNH by adapting it to both CNN and Transformer architectures.

\subsection{Architecture Settings}
For CNN architectures, we use authentic ResNet~\cite{he2016deep} as our backbone model and prototype PNNH to establish plain residual-free deep CNN architectures.

\noindent \textbf{CIFAR-10 \& CIFAR-100.}
On CIFAR-10, we prototype PNNH in the same fashion as ResNet and Wide-ResNet counterparts. 
For example, on ResNet-20, PNNH replaces vanilla residual blocks with a trainable learner and an autoencoder following Eq. \ref{eq:vanilla_coder_func}.
It adopts 3 blocks within each stage for all 3 stages of the network. 
For Wide ResNet~\cite{zagoruyko2016wide}, we use width multiplier and no dropout~\cite{srivastava2014dropout} when applying PNNH to original wide ResNet architectures, such as WRN-28-10~\cite{zagoruyko2016wide}.

\noindent \textbf{ImageNet.} We use ResNet 18/34 as a representative of vanilla residual blocks and prototype PNNH in plain convolutional neural nets with 18/34 layers following Eq. \ref{eq:vanilla_coder_func}. 
For residual bottleneck block designs, we use ResNet-50/ResNet-101 as backbones and apply PNNH to construct 50/101-layer convolutional neural nets following Eq. \ref{eq:rb_coder_func}. 
For inverted residual bottleneck blocks, we use MobileNet-v2 as a backbone and craft plain mobile CNNs following Eq. \ref{eq:irb_coder_func}. 
For Swin Transformer~\cite{liu2021swin}, we adopt the smallest model (i.e., Swin-T) and use Eq. \ref{eq:vanilla_coder_func}. Note that MLP is a special case of the residual block containing 2 3$\times$3 convolution blocks, with a kernel size of 1 for each convolution.

\begin{figure}[b]
\vspace{-2em}
\begin{center}
    \includegraphics[width=0.95\linewidth]{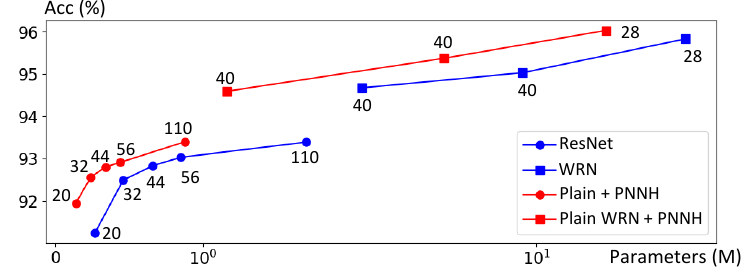}
    \vspace{-1em}
\end{center}
\caption{CIFAR-10 Accuracy-Parameter trade-off on various ConvNets. Number: Network Depth.}
\label{fig:cifar10_traedoff}
\vspace{-1em}
\end{figure}
\subsection{CIFAR-10/CIFAR-100 Results and Analysis}
\label{ref:sec_exp}
We prototype PNNH on the aforementioned models in Section 5.1 and construct plain CNN architectures. We train all CNNs for 300 epochs with an initial learning rate of 0.1, batch size of 128, weight decay of 3e-4, cosine learning rate schedule~\cite{loshchilov2016sgdr} and standard data augmentation.
This recipe gives on-par accuracy for ResNet and Wide-ResNet baselines. 
We use SGD with a momentum of 0.9 to optimize plain CNNs enabled by PNNH, without using dropout~\cite{srivastava2014dropout}.
We compare accuracy-parameter trade-off between PNNH-enabled plain CNNs and ResNets on CIFAR-10 in Figure \ref{fig:cifar10_traedoff}, and compare the best accuracy obtained in the CIFAR-10 / CIFAR-100 test dataset with prior-art and current state-of-the-art plain networks in Table \ref{tab:cifar_table}.  
We compute the mean and standard deviation of PNNH performance over 3 runs.

\noindent \textbf{Parameter Efficiency over ResNets.}
PNNH paradigm wins the accuracy-parameter trade-off over classic hand-crafted ResNets and Wide ResNets.
Thanks to the weight-sharing autoencoder, plain PNNH-enabled neural nets can achieve 2$\times$ parameter efficiency compared to their vanilla ResNet counterparts.
As such, PNNH demonstrates that a plain CNN architecture can achieve good depth scalability without additional engineering efforts.

\begin{table}[t]
    \begin{center}
    \vspace{-1em}
    \caption{Test error comparison on CIFAR-10/CIFAR-100.}
    \scalebox{0.72}{
    \begin{tabular}{|c|c|c|c|}
        \hline
         \textbf{\multirow{2}{*}{Architecture}} & \textbf{C10 Test} & \textbf{C100 Test} & \textbf{Params} \\
          &  \textbf{Err. (\%)} & \textbf{Err. (\%)} & \textbf{(M)} \\
         \hline
         ResNet-110\tiny{}{~\cite{he2016deep}} & 6.61\tiny{}{$\pm$0.16} & 27.22 & 1.7 \\
         \multirow{2}{*}{DiracNet\tiny{}~\cite{zagoruyko2017diracnets}} & 5.16\tiny{}{$\pm$0.14} & - & 9.1 \\
         & 4.75\tiny{}{$\pm$0.16} & 21.54\tiny{}{$\pm$0.18} & 36.5 \\
         ResDistill-18\tiny{}~\cite{li2020residual} & 4.89\tiny{}{$\pm$0.08} & 21.76\tiny{}$\pm$ 0.04 & 11.5 \\
         RepVGG-A2\tiny{}~\cite{ding2021repvgg} & 4.65 & - & 24.1 \\
         \hline
         FractalNet\tiny{}~\cite{larsson2016fractalnet} & 5.22 & 23.30 & 38.6 \\
         RMNet 26$\times$3\_32\tiny{}~\cite{meng2021rmnet} & 4.19 & 20.84 & 8.8 \\
         DenseNet ($k=12$)\tiny{}~\cite{huang2017densely} & 4.10 & 20.20 & 7.0 \\
         WRN-28-10\tiny{}~\cite{zagoruyko2016wide} & 4.0 & 20.50 & 36.5 \\
         \hline
         \textbf{Plain-110} & 36.80 & DIVERGE & 1.7 \\
         \textbf{Plain-110, with PNNH} & 6.37\tiny{}$\pm$0.06 & 28.85\tiny{}$\pm$0.20 & \textbf{0.88} \\
         \textbf{Plain(WRN-28-10), with PNNH} & \textbf{3.90\tiny{}$\pm$0.05} & \textbf{19.83\tiny{}$\pm$0.17} & 18.3 \\
         \hline
    \end{tabular}
    \label{tab:cifar_table}
    }
    \vspace{-2em}
    \end{center}
\end{table}

We further prototype PNNH on wider versions of the ResNet, Wide-ResNet 28-10 (WRN-28-10).
With PNNH, WRN-28-10 achieves better accuracy while saving 2$\times$ parameters compared to the vanilla, Wide ResNet-28-10.
Compared to RepVGG, WRN-28-10 + PNNH achieves a better accuracy with 1.3$\times$ fewer parameters.
The PNNH paradigm allows a plain CNN to achieve on-par accuracy with more complex design such as DenseNet.
Note that without an internal path, plain CNN architectures converge much worse (or even diverges) on CIFAR-10/CIFAR-100.

\noindent \textbf{Preserving Input Improves Scalability.}
PNNH paradigm is crafted based on our understanding of residual learning in plain CNN architectures.
Thus, PNNH shows consistent performance gain with more layers (i.e., 40 vs. 28) and even wider structures (i.e., 10 v.s. 2/4) in the backbone architecture.

\noindent \textbf{Accuracy-latency Trade-off.} 
We compare the performance over various metrics, including training/inference speed and parameter count. 
Table \ref{tab:imagenet_results} demonstrates the key results on ImageNet-1K benchmark.
The training/inference speed is measured on a GTX 1080-Ti using a batch size of 128.
\begin{table*}[t]
    \caption{ImageNet-1K results. All models use 224$\times$224 input size. "$*$": Results collected from ~\cite{li2020residual}.}
    \begin{center}
    \scalebox{0.95}{
    \begin{tabular}{|c|c|c|c|c|c|}
    \hline
    \textbf{\multirow{2}{*}{Model}} & \textbf{Scratch?} & \textbf{Training Speed} & \textbf{Inference Speed} & \textbf{Top-1} & \textbf{Params} \\
    & & \textbf{(Images/sec)} & \textbf{(Images/sec)} & \textbf{Err. (\%)} & \textbf{(M)} \\
    \hline
    Plain-34 & \checkmark & 414 & 1447 & 28.4 & 21.4 \\
    ResNet-18 & \checkmark & 631 & 2334 &  30.2 & 11.7 \\
    \textbf{Plain-18 + PNNH} & \checkmark & 704 & 2533 & 29.9 & 6.4 \\
    ResNet-34 & \checkmark & 389 & 1312 & 26.7 & 21.6 \\
    DiracNet-34 & & - & - & 27.8 & 21.6 \\
    ResDistill-34 & & - & - & 26.2$^{*}$ & 21.6 \\
    \textbf{Plain-34 + PNNH} & \checkmark & 414 & 1447 & 27.5 & 11.4 \\
    \hline
    Plain-50 & \checkmark & 224 & 798 & 30.7$^{*}$ & 22.8 \\
    ResNet-50 & \checkmark & 169 & 629 & 23.9 & 25.5 \\
    \textbf{Plain-50 + PNNH} & \checkmark & 224 &  798 & 24.2 & 15.4 \\
    ResNet-101 & \checkmark & 114 & 409 & 22.7 & 44.5 \\
    \textbf{Plain-101 + PNNH} & \checkmark & 129 & 454 & 22.8 & 23.6 \\ 
    \hline
    MobileNet-V2 & \checkmark & 345 & 1577 & 28.0 & 3.4 \\
    \textbf{Plain-MobileNet-V2 + PNNH} & \checkmark & 446 & 1615 &  28.5 & 3.1 \\
    \hline
    Swin-T & \checkmark &  - & - & 20.0 & 28.3 \\
    \textbf{Plain Swin-T + PNNH} & \checkmark & - & - & 20.3 & 27.3 \\
    \hline
    \end{tabular}
    }
    \end{center}
    \vspace{-2em}
    \label{tab:imagenet_results}
\end{table*}

\subsection{ImageNet-1K Results and Analysis}
We apply the PNNH paradigm to train the large-scale ImageNet classifier on 1.28M ImageNet-1K training images without using additional data. 
To ensure a fair comparison, we prototyped PNNH on plain CNN architectures with 18/34 layers, composed of vanilla residual blocks.
We also prototype PNNH on plain CNNs with 50/101 layers, composed of residual bottleneck blocks, and MobileNet-V2-like plain CNNs composed of inverted residual bottleneck blocks.
For 18/34-layer ResNet-style and MobileNet-style plain neural nets, We preprocess ImageNet training data with Inception preprocessing~\cite{szegedy2017inception}, and train models with an initial learning rate of 0.256, batch size of 256, weight decay of 1e-5, and a cosine learning rate schedule~\cite{loshchilov2016sgdr} with warm-up for 120 epochs.
For 50/101-layer ResNet-style plain neural nets, we enable PNNH and employ the timm ResNet-50 training recipe~\cite{wightman2021resnet}. 
Since the pipeline is optimized on ResNet-50 and far from optimal on PNNH, we compare with the original ResNet-50/ResNet-101 result to demonstrate the ability to train plain neural nets towards residual-level performance. 
For Transformer-style neural nets, we adopt the backbone of Swin-T~\cite{liu2021swin} and employ the Swin-T training recipe from timm~\cite{wightman2021resnet}. 

\begin{figure}[b]
\vspace{-2em}
\begin{center}
    \includegraphics[width=1.0\linewidth]{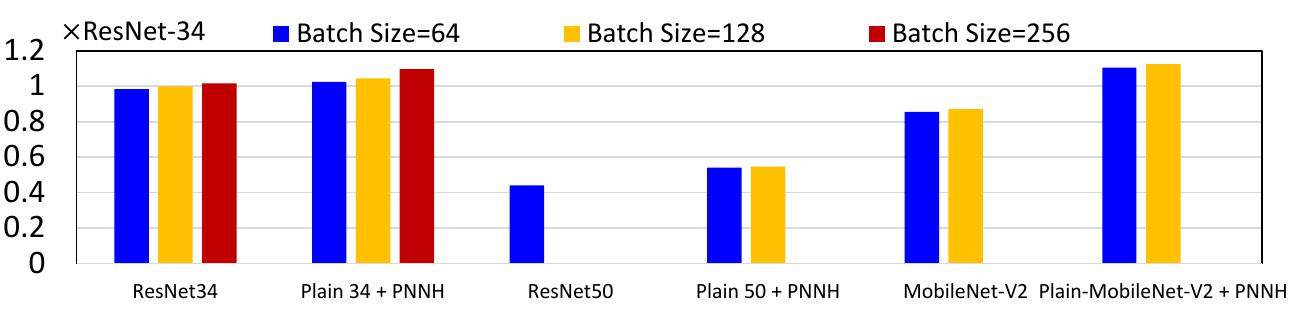}
    \vspace{-2em}
    \caption{Training throughput of plain/residual ConvNets. Missing bar: Out-of-memory on GPU.}
\label{fig:imagenet_eff}
\end{center}
\end{figure}

\noindent \textbf{Training \& Memory Efficiency.}
We illustrate the training throughput of PNNH paradigm relative to ResNet-34 in Figure \ref{fig:imagenet_eff}. Here, all latency number is measured on a NVIDIA GTX 1080 TI with 11GB memory.
With PNNH, we observe up to $\sim 0.1\times$ on vanilla residual blocks, up to $\sim 0.25\times$ on residual bottleneck blocks, and up to $\sim 0.3\times$ on inverted residual bottleneck blocks. 
This strengthen the demonstration of the hardware efficiency in full 3$\times$3 convolutional networks thanks to existing hardware optimizations, especially in inverted residual bottleneck blocks. In addition, the PNNH paradigm has a less expensive memory consumption during training. With a limited 11GB GPU memory, PNNH supports a maximum training batch size of 128, while vanilla ResNet-50 can only support a maximum training batch size of 64. This also justifies the training efficiency of PNNH paradigm.


Our experimental evaluation supports ``The Plain Neural Net Hypothesis'' on a large-scale image classification dataset. On plain neural nets, the 34-layer PNNH-enabled neural net brings up to $\sim$1.0\%/$\sim$5.0\% top-1 accuracy gain, with 2x parameter savings under the original ResNet training pipeline.
Compared to plain CNN architectures trained with engineering techniques such as distillation and reparameterization, a plain 50/101-layer neural net with PNNH enabled can achieve on-par performance with original ResNet 50/101 baselines, while enjoying 10\% higher training efficiency and 2$\times$ parameter efficiency. When transferred to mobile IRB blocks and Transformer MLP residual blocks, plain PNNH-enabled neural nets still maintain competitive performance, achieving an accuracy on par with a similar parameter count. 

\noindent \textbf{Best Scenario for PNNH.} As PNNH saves half of the parameters, it is the best fit to adapt over-parameterized residual convolutional neural nets towards plain convolutional nerual nets, without sacrificing the performance metrics. As a result, PNNH performs on par or better on small benchmarks such as CIFAR-10 / CIFAR-100. In addition, PNNH has a narrower accuracy gap on Inverted Bottleneck blocks, such as MobileNet-v2 IRB blocks and Transformer MLP residual blocks. This is because in IRB/vision Transformer Inverted Bottlenecks, the coder part is much smaller than the learner part, preserving more representation power for the learner part in PNNH.

%% file: _txt/7_Conclusion.tex
\section{Conclusion}
In this work, we peek behind the curtains of residual learning and discover the phenomenon of ``dissipating inputs'' as a result of convergence failure in plain CNNs.
We make a pioneering effort to offer a formal theoretical analysis on residual learning.
Then, we conclude our discoveries with ``The Plain Neural Net Hypothesis'' (PNNH) and propose a novel paradigm to practice PNNH on plain neural nets.
PNNH pioneers the training of deep and scalable plain CNNs from
scratch, and provides better parameter efficiency and training efficiency than existing solutions.
We apply PNNH on various types of plain CNN architectures and plain Transformer MLP architectures. 
Empirical evaluations on CIFAR-10, CIFAR-100, and ImageNet indicate a 0.3$\times$ higher training throughput and 2$\times$ better parameter efficiency compared to ResNet-like ConvNets and/or vision Transformers on vision tasks, with on-par accuracy.

%% file: _txt/8_Appendix.tex
\newpage
\appendix
\onecolumn

\section{Annotations}
\label{sec:annotation}
We list the annotations in this paper for reference in Table \ref{tab:annotations}.
\begin{table}[h]
\vspace{-1.5em}
    \caption{Annotations.}
    \begin{center}
    \begin{tabular}{|c|c|}
    \hline
    \textbf{Symbol or Name} & \textbf{Description}  \\
    \hline  
    $W_{conv}$ & Weight of a convolution layer. \\
    $W_{enc}$ & Weight of an encoder layer. \\
    $W_{dec}$ & Weight of a decoder layer. \\
    $\mu_{x}$ & The mean of input $x$, usually for plain and/or residual layers. \\
    $\sigma^2_{x}$ & The variance of input $x$, usually for plain and/or residual layers. \\
    $\delta$ & The precision of bound for surviving neurons.  \\
    $\epsilon$ & The minimum remaining neuron responses that does not allow ``dissipating inputs''. \\
    $\circledast$ & Convolution. \\
    $ReLU(\cdot)$ & Rectified Linear Unit activation function. \\
    \hline
    Neuron (Responses) & The output of a parameterized layer in CNN, for example, convolution.\\
    Activation & Output of neuron (responses) that pass thorough a non-linearity, for example, ReLU. \\
    \hline
    \end{tabular}
    \end{center}
    \label{tab:annotations}
    \vspace{-1em}
\end{table}

\section{Proof of Theorems}
\label{sec:proofs}
\begin{figure}[t]
\begin{center}
    \includegraphics[width=0.7\linewidth]{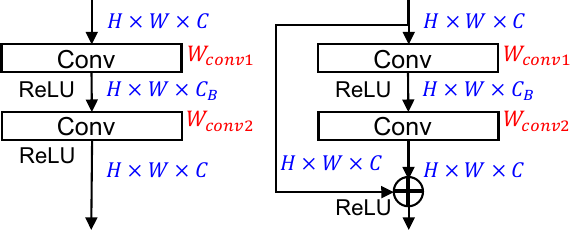}
     \caption{Overview of \textbf{(left)} plain learning and \textbf{(right)} residual learning. We mark activations in \textcolor{blue}{blue} and mark weights in \textcolor{red}{red}.}
    \label{fig:resnet_overview}
    \vspace{-2em}
\end{center}
\end{figure}

Before showing the proofs of theorems, we again emphasize the overview of plain learning and residual learning in Figure \ref{fig:resnet_overview}.
We use the vanilla residual block as an example.

Given an input $x$, we again demonstrate the response $y$ of popular residual learning parameterized by 2 convolutional layers $W_{conv1}$ and $W_{conv2}$ as follows:
\begin{equation}
    y = ReLU(x + ReLU(x \circledast W_{conv1}) \circledast W_{conv2})
\end{equation}

\noindent \textbf{Theorem 3.2}
Given an input $x \in \mathbb{R}^{N \times C \times H \times W}$ with mean $\mu_{x}$ and variance $\sigma_{x}^2$, the randomly initialized weight $W \in \mathbb{R}^{O \times I \times K \times K}$ is independent of the weight.
The weight distribution in He-normal initialization~\cite{he2015delving} ensures: 
\begin{equation}
\mathop{\mathbb{E}}[x \circledast W]=0, Var[x \circledast W] = 2\sigma_{x}^2,
\end{equation}
Here, $K$ denotes 
This is because He-normal adopts a zero-mean weight initialization approach, and by the independence of statistics, it is easy to see that
\begin{equation}
    \mathop{\mathbb{E}}[x \circledast W] = \mathop{\mathbb{E}}[\sum_{i<k, j<k}x_{:, :, i, j}W_{:, :, i, j}]=\sum_{i<k, j<k}\mathop{\mathbb{E}}[x_{:, :, i, j}W_{:, :, i, j}] = 0
    \label{eq:proof3.2.1}
\end{equation}
In Eq.\ref{eq:proof3.2.1}, the last inequality holds because $X$ and $W$ are statistically independent. Thus, Eq. \ref{eq:proof3.2.1} ends up with $\sum_{i<k, j<k}\mathop{\mathbb{E}}[X^{(:, :, i, j)}]\mathop{\mathbb{E}}[W^{(:, :, i, j)}]=0$, as $W$ is initialized from a zero-mean distribution (i.e., He-normal~\cite{he2015delving}).
Notably, He-normal is derived under this assumption and empirically works well to scale the variance of activations.

Hence, the consequent Conv-BN doublet in plain learning generates zero-mean outputs following 2 consecutive convolution layers.
The presence of residual connection yields $\mu_{x}$ for the output response before the final ReLU activation.

Next, we proceed with the variance. 
Given an input with variance $\sigma_{x}^2$, He-normal initialization~\cite{he2015delving} already gives $2\sigma_{x}^2$ for the variance of each convolution layer prior to ReLU activation.
As the weights are symmetrically distributed, He et al. also hypothesizes that ReLU non-linearity cuts off half of the activations, thus leads to an activation with variance $\sigma_{x}^2$.
Notably, the presence of a residual connection adds the original input of variance $\sigma_{x}^2$ before non-linearity, yielding a total of $3\sigma_{x}^2$ variance in the output response before the last ReLU activation.
Notably, the last ReLU activation halfs the variance, yet still fails to achieve a scale-invariant output activation.

\noindent \textbf{Remarks.} The above analysis reveals why 'addition' operation has to be placed before  the 'ReLU' non-linearity: otherwise, zero-mean responses will directly pass the ReLU activation and drop all information as plain layers. The observation that variance may exponentially grow in ResNet activation fits the observation in recent normalizer-free ResNet works~\cite{zhang2019fixup}, where batch normalization layers
are needed to address the exploding variance.

\noindent \textbf{Theorem 3.3}
The key idea lies in bounding a single-sided tail probability given the distribution $x$ centered at $\mathop{\mathbb{E}(x)}=\mu_{x}$, with standard deviation $\sigma_{x}^2$ of inputs.
Here, we first introduce the Chebyshev-Cantelli inequality~\cite{boucheron2013concentration} as follows:

\begin{theorem}
\textbf{(Chebyshev-Cantelli inequality)}
For any $\lambda>0$, given any real-valued distribution $X$ with mean $\mathop{\mathbb{E}}(x)$ and variance $\sigma^2$, the lower tail probability can be upper bounded by:
\begin{equation}
    Pr(X-\mathop{\mathbb{E}}[x] \leq -\lambda) \leq \frac{\sigma^2}{\sigma^2 + \lambda^2}
\end{equation}
\vspace{-1em}
\label{theroem:cantelli}
\end{theorem}
Let $\lambda=\delta=o(\sigma_{x})$ be a small value that determines the precision of the bound.
Simply plug in the mean/variance of output responses of plain layers: $\mathop{\mathbb{E}}[x]=0$, $\sigma_{x}^2$ yields the upper bound that a neuron does not survive in ReLU non-linearity, and thus leads to the upper bound of a neuron surviving ReLU non-linearity in Theorem \ref{eq:plain_layer_bound}. Similarly, we can derive the upper bound of tail probability for residual layers using the statistics of the corresponding responses.

\noindent \textbf{Remarks:} In practice, due to covariate shift and the introduction of bias variable, the output response will no longer be zero-mean, for example, have a mean of $\mu_{y}$.  By plugging the new mean into Chebyshev-Cantelli inequality, the new lower bound is $\frac{(\mu_{y}+\delta)^2}{4\sigma_{x}^2+(\mu_{y}+\delta)^2}$ for plain layers and $\frac{(\mu_{x}+\mu_{y}+\delta)^2}{9\sigma_{x}^2 + {(\mu_{x}+\mu_{y}+\delta)}^2}$ for residual layers, and the aforementioned analysis holds.

Thus, to avoid the 'dissipating input' phenomenon and keep $\epsilon$ surviving neurons, the number of plain layers should satisfy:
\begin{equation}
    (\frac{\delta^2}{4\sigma_{x}^2+\delta^2})^{N_{plain}} < \epsilon 
    \label{eq:plain_bound_layers_process}
\end{equation}
And the number of residual layers should satisfy:
\begin{equation}
    (\frac{(\mu_{x}+\delta)^2}{9\sigma_{x}^2+(\delta+\mu_{x})^2})^{N_{residual}} < \epsilon 
    \label{eq:res_bound_layers_process}
\end{equation}
Solving Eq. \ref{eq:plain_bound_layers_process} and Eq. \ref{eq:res_bound_layers_process} gives the bound in Eq. \ref{eq:plain_layer_bound} and Eq. \ref{eq:residual_layer_bound}.

\section{Choice of Task in Coder}
\label{sec:pretext_study}
As is shown in Eq. \ref{eq:ae_objective}, the coder path selects the rectified normal distribution as the self-supervised task to maintain input information over rectified activations.
Here, we study the effects of various self-supervision tasks on training the coder path as follows.

\noindent \textbf{No Coder.} No coder is deployed in deep plain CNN.

\noindent \textbf{No Task.} The coder is randomly initialized with He-normal distribution as convolution parts.

\noindent \textbf{Normal.} The coder is trained with data sampled from a normal distribution.

\noindent \textbf{Uniform.} The coder AE is trained with data sampled from a uniform distribution from $\mathrm{U}(0, 1)$.

\noindent \textbf{Identity.} The coder adopts an identity mapping in Eq. \ref{eq:aeinit}. This randomly drops channels from input.
\begin{table}[h]
\begin{center}
\caption{Effects of coder pretext task on ResNet-110. We show the average accuracy over 3 runs.}
\begin{tabular}{|c|c|c|}
    \hline
         \textbf{Pretext Task} & \textbf{Accuracy (\%)} & \textbf{Change ($\Delta$)} \\
    \hline
    \textbf{Rectified Normal} & \textbf{93.63} & \textbf{0.00} \\
    No Coder & DIVERGE & - \\
    No Task & 85.21 & -8.42 \\
    Normal & 88.37 & -5.26 \\
    Uniform & 93.15 & -0.48 \\
    Identity & 93.22 & -0.41\\
    \hline
    \end{tabular}
\label{tab:pretext_study}
\end{center}
\end{table}

Table \ref{tab:pretext_study} shows the effects of various pretext tasks by coders on ResNet-110, using the PNHH paradigm.
Without a coder structure, the derived 110-layer Plain CNN loses all information on inputs following the aforementioned analysis and diverges at early epochs.
Notably, a pretext task that accurately maintains the nonnegative inputs (i.e. \textbf{ Rectified Normal} and \textbf{Uniform}) gives the coder best generalization, and \textbf{Rectified Normal} provides the most accurate priors on inputs.
This is because the inputs to the coder are usually sub-Gaussian non-negative activations. More importantly, \textbf{ Identity} establishes the baseline for residual learning in both ResNet~\cite{he2016deep} and RMNet~\cite{meng2021rmnet}.
The results show that as one of the critical solutions in the coder, the identity mapping within a residual connection has a very competitive performance, yet the coder function justifies that there exists a better solution than pure residual learning through the identity matrix by edging 0.18\% accuracy.

\section{Choice of BN in Learner.}
\label{sec:choiceBN}
\begin{figure}[t]
\begin{center}
    \includegraphics[width=0.9\linewidth]{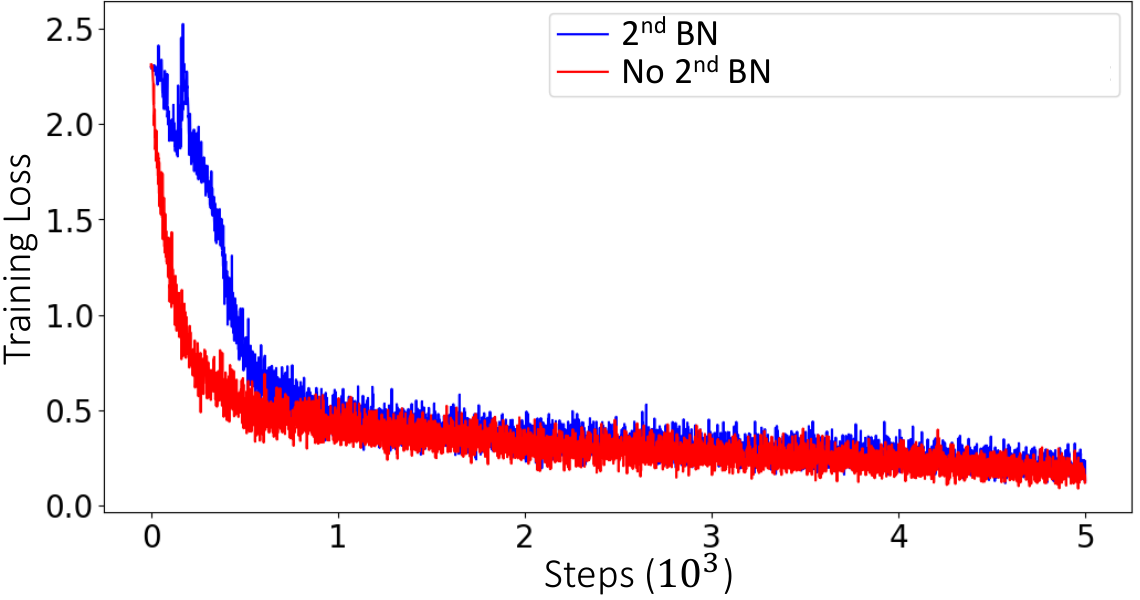}
    \caption{Learning dynamics of ResNet-110 at initial 5K steps with/without first BN layers, using PNHH paradigm. 
    Removing the $2^{nd}$ BN is helpful for obtaining faster convergence and achieving better generalization.}
    \label{fig:convcoder_bn}
\end{center}
\end{figure}

The use of normalization layers needs careful consideration in the learner: arbitrarily stacking BN layers may compromise the input without the explicit use of a residual connection.
We observe that the $2^{nd}$ BN layer may induce a significant accuracy loss in PNHH, see Figure \ref{fig:convcoder_bn}.
This is because BN drops the mean of activations back to zero before ReLU, thus forcing $\mu_{x}=0$ in Eq. \ref{eq:residual_layer_bound}.
As such, the lower bounds in Eq. \ref{eq:residual_layer_bound} is no longer better than the plain counterparts in Eq. \ref{eq:plain_layer_bound} and leads to more information loss on inputs.
Thus, we drop the $2^{nd}$ BN layer in the learner to strive for higher accuracy and benefit from better training efficiency and memory reduction.

Under the PNHH paradigm, a 110-layer plain CNN derived from ResNet-110 ends up with 93. 20\% Testing accuracy in the CIFAR-10 dataset with $2^{nd}$ batch normalization.
The accuracy further improves to 93. 63\% testing accuracy in the CIFAR-10 data set
while we remove $2^{nd}$ BN. 

